\documentclass[12pt]{article}
\usepackage{amsmath}
\usepackage{amsfonts}
\usepackage{mathtools}
\usepackage{subcaption}
\usepackage[margin=1in]{geometry}
\usepackage{array}
\usepackage{bbm}
\usepackage{graphicx}
\usepackage{footnote}
\usepackage{amssymb,amsthm,amsfonts}

\usepackage{caption}
\usepackage{algorithm,algpseudocode}
\usepackage{booktabs}

\usepackage{hyperref}

\title{\textbf{Learning Graph-Level Representations with Recurrent Neural Networks}}

\author{ Yu Jin, Joseph F. JaJa\\
	Department of Electrical and Computer Engineering\\
	Institute for Advanced Computer Studies\\
	University of Maryland, College Park \\
	Email: \href{mailto:yuj@umd.edu}yuj@umd.edu, \href{mailto:joseph@umiacs.umd.edu} joseph@umiacs.umd.edu
}

\begin{document}
\maketitle
\begin{abstract}
	Recently a variety of methods have been developed to encode graphs into low-dimensional vectors that can be easily exploited by machine learning algorithms. The majority of these methods start by embedding the graph nodes into a low-dimensional vector space, followed by using some scheme to aggregate the node embeddings. In this work, we develop a new approach to learn graph-level representations, which includes a combination of unsupervised and supervised learning components. We start by learning a set of node representations in an unsupervised fashion. Graph nodes are mapped into node sequences sampled from random walk approaches approximated by the Gumbel-Softmax distribution. Recurrent neural network (RNN) units are modified to accommodate both the node representations as well as their neighborhood information. Experiments on standard graph classification benchmarks demonstrate that our proposed  approach achieves superior or comparable performance relative to the state-of-the-art algorithms in terms of convergence speed and classification accuracy. We further illustrate the effectiveness of the different components used by our approach.
\end{abstract}

\noindent Graphs are extensively used to capture relationships and interactions between different entities in many domains such as  social science, biology, neuroscience, communication networks, to name a few.  Machine learning on graphs has recently emerged as a powerful tool to solve graph related tasks in various applications such as recommendation systems, quantum chemistry, and genomics \cite{hamilton2017representation,bronstein2016geometric,duvenaud2015convolutional,dai2016discriminative,gilmer2017neural,kipf2016semi}.  Excellent reviews of recent machine learning algorithms on graphs appear in \cite{hamilton2017representation,khasanova2017graph}. However, in spite of the considerable progress achieved, deriving graph-level features that can be used by machine learning algorithms remains a challenging problem, especially for networks with complex substructures.

\begin{figure*}[t!]
	\centering
	\includegraphics[width=0.9\linewidth]{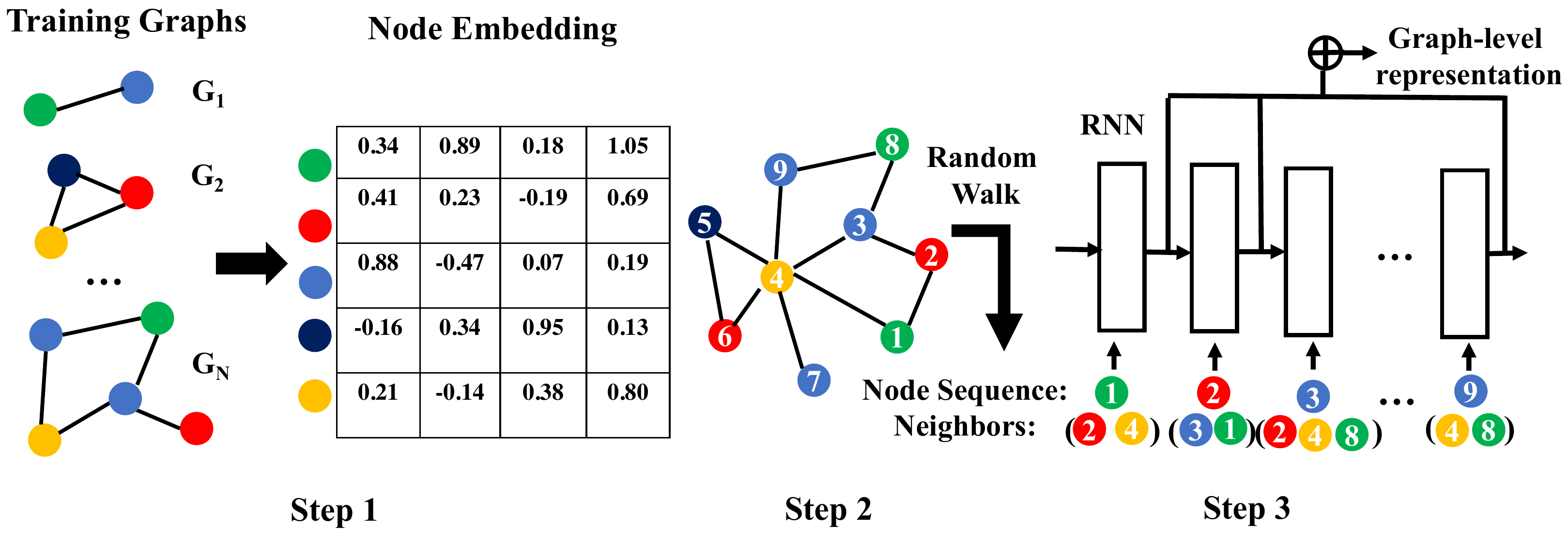}
	\caption{Graph recurrent neural network model to learn graph-level representations. Step 1: Node embeddings are learned from the graph structures and node features over the entire training samples. Step 2: Graph node sequences are continuously sampled from a random walk method approximated by the Gumbel-Softmax distribution. Step 3: Node embeddings as well as the node sequences are used as input by a modified RNN model to learn graph-level features to predict graph labels. Step 2 and 3 form a differentiable supervised learning model with both both random walk and RNN parameters learned from back-propagation with respect to the classification objective.}
	\label{overview}
\end{figure*}

In this work, we develop a new approach to learn graph-level representations of variable-size graphs based on the use of recurrent neural networks (RNN). RNN models with gated RNN units such as Long Short Term Memory (LSTM) have outperformed other existing deep networks in many applications. These models have been shown to have the ability to encode variable-size sequences of inputs into group-level representations; and to learn long-term dependencies among the sequence units. The key steps of our approach include a new scheme to embed the graph nodes into a low dimensional vector space, and a random walk based algorithm to map graph nodes into node sequences approximated by the Gumbel-Softmax distribution. More specifically, inspired by the continuous bag-of-word (CBOW) model for learning word representation \cite{mikolov2013efficient,mikolov2013distributed}, we learn the node representations based on the node features as well as the structural graph information relative to the node. We subsequently use a random walk approach combined with the Gumbel-Softmax distribution to continuously sample graph node sequences where the parameters are learned from the classification objective. The node embeddings as well as the node sequences are used as input by a modified RNN model to learn graph-level features to predict graph labels. We make explicit modifications to the architectures of the RNN models to accommodate inputs from both the node representations as well as its neighborhood information. We note that the node embeddings are trained in an unsupervised fashion where the graph structures and node features are used. The sampling of node sequences and RNN models form a differentiable supervised learning model to predict the graph labels with parameters learned from back-propagation with respect to the classification objective. The overall approach is illustrated in Figure \ref{overview}. 

Our model is able to capture both the local information from the dedicated pretrained node embeddings and the long-range dependencies between the nodes captured by the RNN units. Hence the approach combines the advantages of previously proposed graph kernel methods as well as graph neural network methods, and exhibits strong representative power for variable-sized graphs. 

The main contributions of our work can be summarized as follows,
\begin{itemize}
	\item \textbf{Graph recurrent neural network model.} We extend the RNN models to learn graph-level representations from samples of variable-sized graphs.
	\item \textbf{Node embedding method.} We propose a new method to learn node representations that encode both the node features and graph structural information tailored to the specific application domain.
	\item \textbf{Map graph nodes to sequences.} We propose a parameterized random walk approach with the Gumbel-Softmax distribution to continuously sample graph nodes sequences with parameters learned from the classification objective.
	\item \textbf{Experimental results.} The new model achieves improved performance in terms of the classification accuracy and convergence rate compared with the state-of-the art methods in graph classification tasks over a range of well-known benchmarks.
\end{itemize}

\section{Related Work}
\subsubsection{Graph kernels} Graph kernel methods are commonly used to compare graphs and derive graph-level representations. The graph kernel function defines a positive semi-definite similarity measure between arbitrary-sized graphs, which implicitly corresponds to a set of representations of graphs. Popular graph kernels encode graph structural information such as the numbers of elementary graph structures including walks, paths, subtrees with emphasis on different graph substructures \cite{gartner2003graph,kashima2003marginalized,horvath2004cyclic}. Kernel based learning algorithms such as Support Vector Machine (SVM) are subsequently applied using the pair-wise similarity function to carry out specific learning tasks. The success of the graph kernel methods relies on the design of the graph kernel functions, which often depend on the particular application and task under consideration. Note that most of the hand-crafted graph kernels are prespecified prior to the machine learning task, and hence the corresponding feature representations are not learned with the classification objective \cite{douglas2011weisfeiler}. 

\subsubsection{Graph convolutional neural networks}
Motivated by the success of convolutional neural networks (CNN), recent work has adopted the CNN framework to learn graph representations in a number of applications \cite{duvenaud2015convolutional,defferrard2016convolutional,niepert2016learning,kipf2016semi,gilmer2017neural}. The key challenge behind graph CNN models is the generalization of convolution and pooling operators on the irregular graph domain. The recently proposed convolutional operators were mostly based on iteratively aggregating information from neighboring nodes and the graph-level representations are accumulated from the aggregated node embeddings through simple schemes or graph coarsening approaches \cite{kipf2016semi,li2015gated,defferrard2016convolutional}. 

\subsubsection{Recurrent neural networks on graphs}
Recurrent neural network (RNN) models have been successful in handling many applications, including sequence modeling applications thereby achieving considerable  success in a number of challenging applications such as machine translation, sentence classification \cite{cho2014learning}. Our model is directly inspired by the application of RNN models in sentence classification problems, which use RNN units to encode a sequence of word representations into group-level representations with parameters learned from the specific tasks \cite{yin2017comparative}. 

Several recent works have adapted the RNN model on graph-structured data. Li et al. modified the graph neural networks with Gated Recurrent Units (GRU) and proposed a gated graph neural network model to learn node representations \cite{li2015gated}.  The model includes a GRU-like propagation scheme that simultaneously updates each node's hidden state absorbing information from neighboring nodes.  Jain et al. and  Yuan et al. applied the RNN model to analyze temporal-spatial graphs with computer vision applications where the recurrent units are primarily used to capture the temporal dependencies \cite{jain2016structural,yuan2017temporal}. You et al. proposed to use RNN models to generate synthetic graphs \cite{you2018graphrnn} trained on Breadth-First-Search (BFS) graph node sequences. 

\section{Preliminaries}
A graph is represented as the tuple $G = (V, H, A, l)$ where $V$ is the set of all graph nodes with $|V| = n$. $H$ represents the node attributes such that each node  attribute is a discrete element from the alphabet $\Sigma$,  where $|\Sigma| = k$. We assume that the node attributes are encoded with one-hot vectors indicating the node label type, and hence $H\in \mathbb{R}^{n\times k}$. $H_i \in \mathbb{R}^{k}$ denotes the one-hot vector for the individual node $i$ corresponding to the $i^{\text{th}}$. The matrix $A \in \mathbb{R}^{n \times n}$ is the adjacent matrix. From the adjacent matrix, we denote $\mathcal{N}_s(i)$ as the set of neighbors of node $i$ with distance $s$ from node $i$. $l$ is the discrete graph label from a set of graph labels $\Sigma_g$.

In this work, we consider the \emph{graph classification problem} where the training samples are labeled graphs with different sizes. The goal is to learn graph level features and train a classification model that can achieve the best possible accuracy efficiently.

\section{Proposed Approach}
\subsection{Learning node embedding from graph structures}
The node embeddings are learned from the graph structures and node attributes over the entire training graphs. The main purpose is to learn a set of domain-specific node representations that encode both the node features and  graph structural information.

Inspired by the \emph{continuous Bag-of-Words} (CBOW) model to learn word representations, we propose a node embedding model with the goal to predict the central node labels from the embeddings of the surrounding neighbors \cite{mikolov2013efficient,mikolov2013distributed}. Specifically, we want to learn the embedding matrix $E \in \mathbb{R}^{k \times d}$ such that each node  $i$ is mapped to a $d$-dimensional vector $e_i$ computed as $e_i = H_iE$ , and the weight vector $w \in \mathbb{R}^{K}$ representing the weights associated with the set of neighbor nodes $\mathcal{N}_1, \mathcal{N}_2, ..., \mathcal{N}_K$ corresponding to different distances.

The predictive model for any node $i$ is abstracted as follows:

\begin{equation}
	Y_i = f(\sum_{s = 1}^K (w_s \sum_{j\in \mathcal{N}_s(i)}H_jE))
	\label{pretrain_node}
\end{equation}

Each term $w_s\sum_{j\in \mathcal{N}_s(i)}H_jE$ corresponds to the sum of node embeddings from the set of neighbors that are $s$-distance to the center node $i$. $f(\cdot)$ is a differentiable predictive function and $Y_i \in \mathbb{R}^k$ corresponds to the predicted probability of the node type. In the experiment, we use a two-layer neural network model as the predictive function: 
\begin{equation}
	Y_i = \text{Softmax}(W_2\text{ReLU}(W_1F + b_1)+b_2)
\end{equation}
where $F = \sum_{s = 1}^K (w_s \sum_{j\in \mathcal{N}_s(i)}h_jE)$. The loss function is defined as the sum of the cross entropy error over all nodes in the training graphs,

\begin{equation}
	\mathcal{L} = -\sum_{m = 1}^N\sum_{i \in V_m} H_i \ln Y_i
\end{equation}

\subsubsection{Connection with existing node embedding models}
Previous work has described a number of node embedding methods to solve problems in various applications such as recommendation system and link prediction \cite{grover2016node2vec,hamilton2017inductive}. 

Perozzi et al. and Grover et al. proposed \emph{DeepWalk} and \emph{node2vec} which use neighborhood information to derive the node embeddings inspired from Skip-Gram language models \cite{perozzi2014deepwalk,grover2016node2vec}. Their objective is to preserve the similarity between nodes in the original network, which is obviously different from our proposed method. Our method has a similar formulation with Graph Convolutional Network (GCN) and GraphSAGE but the main difference is that they explicitly include the central node embedding aggregated with neighboring nodes to predict the central node labels but our pretrained model only uses the neighbors of the node information \cite{kipf2016semi,hamilton2017inductive}. In addition, their goal is to predict the node label for the unseen nodes in the network while ours is to learn a set of node representations that encode both node and structural information. 

\begin{figure*}[t!]
	\centering
	\includegraphics[width=0.9 \linewidth]{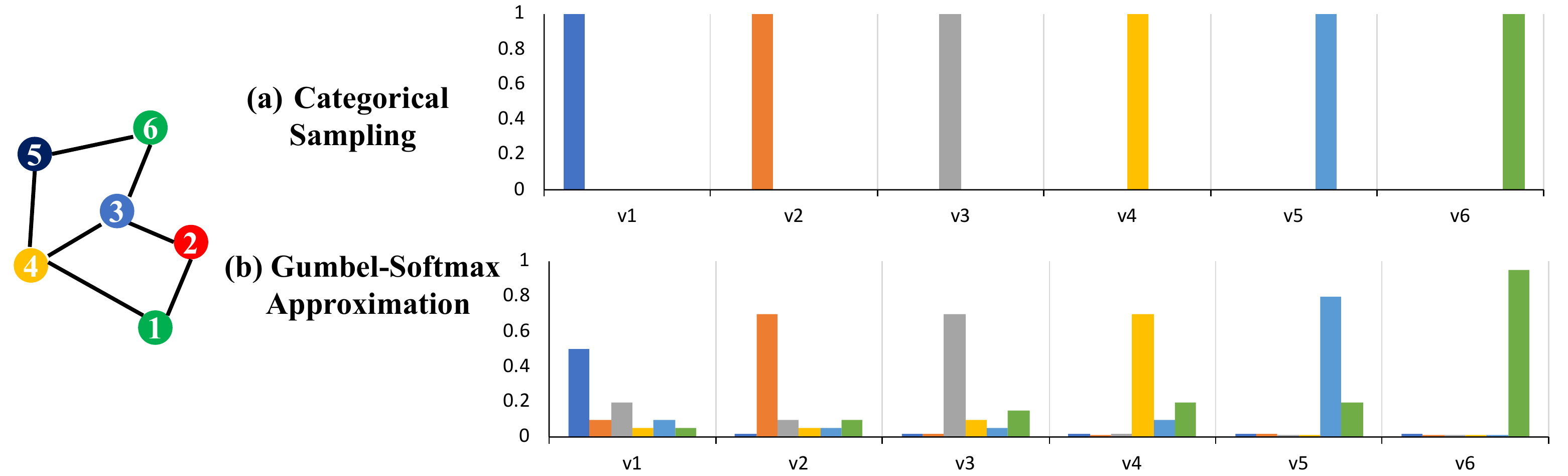}
	\caption{Example of the graph node sequences sampled from the random walk approach. (a) the node sequence consists of categorical samples from the random walk distribution (b) the Gumbel-Softmax approximation of the node sequence.}
	\label{node_sequence}
\end{figure*}

\subsection{Learning graph node sequences}
Since the next state of RNNs depends on the current input as well as the current state, the order of the data being processed is critical during the training task \cite{vinyals2015order}. Therefore it is important to determine an appropriate ordering of the graph nodes whose embeddings, as determined by the first stage, will be processed by the RNN.

As graphs with $n$ nodes can be arranged into $n!$ different sequences, it is intractable to enumerate all possible orderings. Hence we consider a random walk approach combined with the Gumbel-Softmax distribution to generate continuous samples of graph node sequences with parameters to be learned with the classification objective. 

We begin by introducing the weight matrix $W \in \mathbb{R}^{n \times n}$ with parameters $C \in \mathbb{R}^{K_{\text{rw}}}$ and $\epsilon$ defined as follows,

\begin{equation}
	W_{ij} = \begin{cases}
		C_s \text{ if } j \in \mathcal{N}_s(i), s = 1, ..., K_{\text{rw}} \\
		\epsilon \text{ otherwise}
	\end{cases}
\end{equation}

In other words, $W$ is parameterized by assigning the value $C_s$ between nodes with distance $s$ for $s = 1, ..., K_\text{rw}$ and $\epsilon$ for nodes with distance beyond $K_\text{rw}$. The random walk transition matrix $P \in \mathbb{R}^{n\times n}$ is defined as the softmax function over the rows of the weight matrix,

\begin{equation}
	\label{transition}
	P_{ij} = \frac{\exp{W_{ij}}}{\sum_{k=1}^n \exp{W_{ik}}}
\end{equation}

In the following, we use $P_i$ and $W_i$ to  denote the vectors corresponding to $i^\text{th}$ rows of the matrices $P$ and $W$ respectively. The notation $P_{ij}$ and $W_{ij}$ correspond to the matrix elements.

The graph sequence denoted as $S(G) = (v_{\pi(1)}, ..., v_{\pi(n)})$ consists of consecutive graph nodes sampled from the transition probability as in Equation \ref{sample}. $\pi{(i)}$ indicates the node index in the $i^{\text{th}}$ spot in the sequence, and $(v_{\pi(1)}, ..., v_(\pi_{(n)}))$ forms the permutation of $(v_1, ..., v_n)$. Each of the node $v \in \mathbb{R}^{n}$ corresponds to a one-hot vector with $1$ at the selected node index. 

\begin{equation}
	\label{sample}
	v_{\pi(i)} = \text{Sample}(P_{\pi(i-1)})
\end{equation}

Note that the first (root) node is sampled uniformly over all of the graph nodes. 

However, sampling categorical variables directly from the random walk probabilities suffers from two major problems: 1) the sampling process is not inherently differentiable with respect to the distribution parameters; 2) the node sequence may repeatedly visit the same nodes multiple times while not include other unvisited nodes.

To address the first problem, we introduce the \emph{Gumbel-Softmax} distribution to approximate samples from a categorical distribution\cite{jang2016categorical,maddison2016concrete}. Considering the sampling of $v_{\pi(i)}$ from probability $P_{\pi(i-1)}$ in Equation \ref{sample}, the Gumbel-Max provides the following way to draw samples from the random walk probability as

\begin{equation}
	v_{\pi(i)} = \text{one\_hot}(\arg_j\max(g_j + \log P_{\pi(i-1)j}))
\end{equation}

where $g_j$'s are i.i.d. samples drawn from $\text{Gumbel}(0, 1)$ distribution\footnote{The $\text{Gumbel}(0, 1)$ distribution can be sampled with inverse transform sampling by first drawing $u$ from the uniform distribution $\text{Uniform}(0,1)$ and computing the sample $g = -\log(-\log(u))$}. We further use softmax function as a continuous and differentiable approximation to  $\arg\max$. The approximate sample is computed as,

\begin{equation}
	\tilde{v}_{\pi(i)j} = \frac{\exp{((g_j + \log P_{\pi(i-1)j})/\tau)}}{\sum_{k=1}^n \exp{((g_k + \log P_{\pi(i-1)k})/\tau)}}
\end{equation}

\begin{algorithm}[t!]
	\caption{Random Walk Algorithm to Sample Node Sequences with the Gumbel-Softmax Distribution}
	\label{random_walk}
	\begin{algorithmic}[1]
		\State \textbf{Input:} $G = (V, H, A, l)$, the set of neighbors for each node $\mathcal{N}_s(\cdot)$, parameters $C \in \mathbb{R}^K$ and $\epsilon$.
		\State \textbf{Output:} the node sequences, $S(G) = (v_{\pi(1)}, v_{\pi(2)}, ..., v_{\pi(n)})$   
		\State Initialize $W \in \mathbb{R}^{n \times n}$ with $W_{ij} = \begin{cases}
		C_k \text{ if } j \in \mathcal{N}_k(i) \\
		\epsilon \text{ otherwise}
		\end{cases}$
		\State Initialize $T_j(0) = 1$ and $P_{\pi(0)j} = \frac{1}{n}$ for $j = 1 ,...,n$
		
		\For {$i = 1, ..., n$}
		\State $v_{\pi(i)} = \text{Gumbel\_Softmax}(P_{\pi(i-1)})$
		\State $T(i) = T(i-1) \odot (1-v_{\pi(i)})$
		\State $W_{\pi(i)}$ = $v_{\pi(i)} \cdot W$
		\State $P_{\pi(i)} = \text{Softmax}(W_{\pi(i-1)} \odot T(i-1)) $
		\EndFor
	\end{algorithmic}
\end{algorithm}

The softmax temperature $\tau$ controls the closeness between the samples from the Gumbel-Softmax distribution and the one-hot representation. As $\tau$ approaches 0, the sample becomes identical to the one-hot samples from the categorical distribution \cite{jang2016categorical}. 

To address the second problem, we introduce an additional vector $T \in \mathbb{R}^{n}$, which indicates the status of whether specific nodes have been visited, to regularize the transition weight and the corresponding transition probability. Specifically, $T$ is initialized to be all $\mathbf{1}$'s and the next node sequence $v_{\pi(i)}$ is sampled from the following equations where $\odot$ represents element-wise multiplication.,

\begin{gather}
	P_{\pi(i-1)} = \text{Softmax}(W_{\pi(i-1)} \odot T(i-1))\\
	v_{\pi(i)} = \text{Sample}(P_{\pi(i-1)})\\
	T(i) = T(i-1) \odot (1-v_{\pi(i)})
\end{gather}

The introduction of the additional vector $T$ reduces the probability of nodes which have been visited while the Gumbel-Softmax distributions still remain differentiable with respect to the parameters. 

The full algorithm of mapping graph nodes into sequences is described in Algorithm \ref{random_walk} and an example is depicted in Figure \ref{node_sequence}.

\subsection{Recurrent neural network model on graphs}
We adapt the recurrent neural network model especially LSTM to accommodate both the node attributes and the neighborhood information with the node sequences sampled from the random walk approach. As each element $v_{\pi(i)}$ in the node sequence corresponds to a softmax over all the graph nodes, the input node feature denoted as $e_{v_{\pi(i)}}$ and the neighborhood feature denoted as $Nb_{v_{\pi(i)}}$ are computed as the weighted sum of the corresponding node and neighbor embeddings,
\begin{gather*}
	e_{v_{\pi(i)}} = \sum_{j = 1}^n(v_{\pi(i)j} \cdot e_j) \\
	Nb_{v_{\pi(i)}} = \sum_{j = 1}^n(v_{\pi(i)j} \cdot Nb_j)
\end{gather*}

where $e_i$ is the representation of a node as generated by the first stage algorithm and $Nb_i = \sum_{j\in \mathcal{N}_1(t)} e_j$ as the aggregated neighborhood embeddings of node $i$. Given the state of the recurrent units defined by $h_{t+1} = g(h_t, x_t)$, we modify the state update as $h_{t+1} = g'(h_t, x_t, Nb_t)$ to take into account both the node and neighborhood information. The graph-level representation is formed as the sum of hidden units over all the sequence steps as follows.
\begin{equation}
	h_g = \sum_{t = 1}^n h_i
	\label{graph_representation}
\end{equation}

\begin{figure}
	\centering
	\includegraphics[width=0.8\columnwidth]{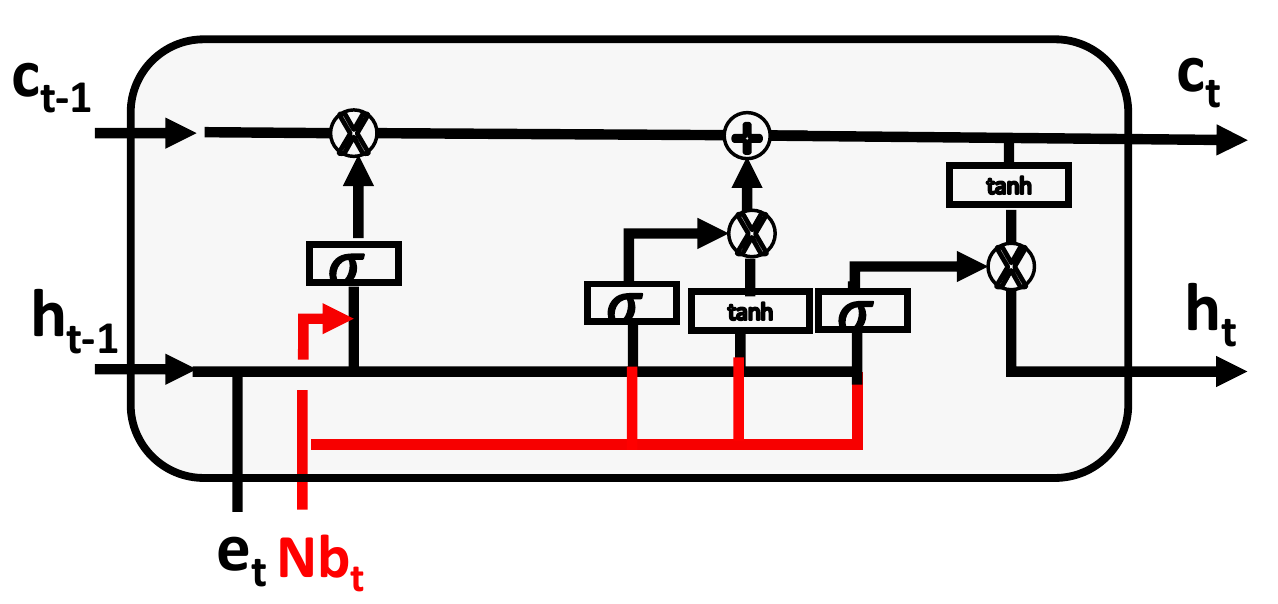}
	\caption{Architecture of graph LSTM model with modified parts highlighted in red.}
	\label{lstm}
\end{figure}

For the LSTM model, we propagate the neighbor information to all the LSTM gates which allows the neighborhood information to be integrated into the gate state as shown in Figure \ref{lstm}. The formulations are as follows,
\begin{gather*}
	i_t = \sigma(W_{ii}e_{v_{\pi(t)}} + \mathbf{W_{ni} Nb_{v_{\pi(t)}}} + W_{hi}h_{t-1} + b_{i}) \\
	f_t = \sigma(W_{if}e_{v_{\pi(t)}} + \mathbf{W_{nf} Nb_{v_{\pi(t)}}}+ W_{hi}h_{t-1} + b_{f}) \\
	g_t = \text{tanh}(W_{ig}e_{v_{\pi(t)}} + \mathbf{W_{ng} Nb_{v_{\pi(t)}}} + W_{hg}h_{t-1} + b_{g}) \\
	o_t = \sigma(W_{io}e_{v_{\pi(t)}} + \mathbf{W_{no} Nb_{v_{\pi(t)}}}  +  W_{ho}h_{t-1} + b_{o})\\
	c_t = f_tc_{t-1} + i_tg_t \\
	h_t = o_t \text{tanh}(c_t)
\end{gather*}

Other RNN architectures such as Gated Recurrent Units (GRU) can be similarly defined by propagating the neighbor information into the corresponding gates. 

\subsection{Discriminative training}
A predictive model is appended on the graph-level representations to predict the graph label. In the experiment, we use a two-layer fully-connected neural network for discriminative training.  All the parameters of recurrent neural networks, the random walk defined above as well as the two-layer neural network predictive models are learned with the back-propagation from the loss function defined as the cross entropy error between the predicted label and the true graph label as in Equation \ref{loss}.

\begin{equation}
	\label{loss}
	\mathcal{L}_g = -\sum_{m = 1}^N l_m\ln y_m
\end{equation}

\begin{table*}[h!]
	\centering
	\resizebox{1.0\textwidth}{!}{%
		\begin{tabular}{cccccccc}
			\toprule
			Datasets & sample size & average $|V|$ &  average $|E|$ & max $|V|$ & max $|E|$ & node labels & graph classes \\
			\midrule
			MUTAG  & 188 & 17.93 & 19.79 & 28 & 33& 7 & 2 \\
			\cmidrule{1-8}
			ENZYMES & 600 & 32.63 & 62.14 & 126 & 149 & 3 & 6\\
			\cmidrule{1-8}
			NCI1 & 4110 & 29.87 & 32.3 & 111 & 119 & 37 & 2\\
			\cmidrule{1-8}
			NCI109 & 4127 & 29.68 & 32.13 & 111 & 119 & 38 & 2\\
			\cmidrule{1-8}
			DD & 1178 & 284.32 & 715.66 & 5748 & 14267 & 82 & 2\\
			\bottomrule
		\end{tabular}
	}
	\caption{Statistics of the graph benchmark datasets \cite{sugiyama2015halting}.}
	\label{dataset}
\end{table*}

\begin{table*}[!ht]
	\centering
	\resizebox{1.0\textwidth}{!}{%
		\begin{tabular}{ccccccccc}
			\toprule
			Datasets & WL subtree & WL edge & WL sp & PSCN & DE-MF & DE-LBP & DGCNN & GraphLSTM \\
			\midrule
			MUTAG & 82.05 & 81.06 & 83.78 &  92.63 & 87.78  & 88.28 & 85.83 &\textbf{93.89}\\
			\midrule
			ENZYMES & 52.22 & 53.17 & 59.05 & N\/A & 60.17 & 61.10 & N\/A &\textbf{65.33}\\
			\midrule
			NCI1 & 82.19 & 84.37 & \textbf{84.55} & 78.59 & 83.04 & 83.72 & 74.44 & 82.73\\
			\midrule
			NCI109 & 82.46 & \textbf{84.49} & 83.53 & 78.59 & 82.05 & 82.16 & N\/A & 82.04\\
			\midrule
			DD & 79.78 & 77.95 & 79.43 & 77.12 & 80.68 & 82.22 & 79.37 & \textbf{84.90}\\
			\bottomrule
		\end{tabular}
	}
	\caption{10-fold cross validation accuracy on graph classification benchmark datasets.}
	\label{classification}
\end{table*}

\subsection{Discussion on isomorphic graphs}
Most previous methods on extracting graph-level representations are designed to be invariant relative to isomorphic graphs, i.e. the graph-level representations remain the same for isomorphic graphs under different node permutations. Therefore the common practice to learn graph-level representations usually involves applying \emph{associative} operators, such as sum, max-pooling, on the individual node embeddings \cite{dai2016discriminative,duvenaud2015convolutional}. 

However as we observe that for node-labeled graphs, isomorphic graphs do not necessarily share the same properties. For example, graphs representing chiral molecules are isomorphic but do not necessarily have the same properties, and hence may belong to different classes \cite{liang1994classification}. In addition, the restrictions on the associative operators limit the possibility to aggregate effective and rich information from the individual units. Recent work, such as the recently proposed GraphSAGE model, has shown that non-associative operators such as LSTM on random sequences performs better than some of the associative operators such as max-pooling aggregator at node prediction tasks \cite{hamilton2017representation}.

Our models rely on RNN units to capture the long term dependencies and to generate fixed-length representations for variable-sized graphs. Due to the sequential nature of RNN, the model may generate different representations under different node sequences. However with the random walk based node sequence strategy, our model will learn the parameters of the random walk approach that generates the node sequences to optimize the classification objective. The experimental results in the next section will show that our model achieves comparable or better results on benchmark graph datasets than previous methods including graph kernel methods as well as the graph neural network methods.

\section{Experimental Results}
We evaluate our model against the best known algorithms on standard graph classification benchmarks. We also provide a discussion on the roles played by the different components in our model. All the experiments are run on  Intel Xeon CPU and two Nvidia Tesla K80 GPUs. The codes will be publicly available.

\subsection{Graph classification}
\subsubsection{Datasets}
The graph classification benchmarks contain five standard datasets from biological applications, which are commonly used to evaluate graph classification methods \cite{shervashidze2011weisfeiler,dai2016discriminative,sugiyama2015halting}. MUTAG, NCI1 and NCI109 are chemical compounds datasets while ENZYMES and DD are protein datasets \cite{wale2008comparison,borgwardt2005shortest}. Relevant information about these datasets is shown in Table \ref{dataset}. We use the same dataset setting as in Shervashidze et al. and Dai et al. \cite{shervashidze2011weisfeiler,dai2016discriminative}. Each dataset is split into 10 folds and the classification accuracy is computed as the average of 10-fold cross validation.

\subsubsection{Model Configuration}
In the pretrained node representation model, We use $K = 2$ with the number of epochs set to 100. Following the standard practice, we use ADAM optimizer with initial learning rate as 0.0001 with adaptive decay \cite{kingma2014adam}. The embedding size $d$ is selected from \{16, 32, 64, 128\} tuned for the optimal classification performance. 

For the random walk approach to sample node sequences, we choose the softmax temperature $\tau$ from $ \{0.5, 0.01, 0.0001\}$ and $K_{\text{rw}} = 2$. The parameters of $C$ are initialized uniformly from $0$ to $5$ and $\epsilon$ is uniformly sampled from $-1$ to $1$.  

For the graph RNN classification model, we use the LSTM units as the recurrent units to process the node sequences which we term as \textbf{GraphLSTM}. The dimension of the hidden unit in the LSTM is selected from \{16, 32, 64\}. A two-layer neural network model is used as the classification model for the graph labels. The dimension of the hidden unit in the classification model is selected from 
\{16, 32, 64\}. ADAM is used for optimization with the initial learning rate as $0.0001$ with adaptive decay \cite{kingma2014adam}.

The baseline algorithms are Weisfeiler-Lehman (WL) graph kernels including subtree kernel, edge kernel and shortest path (sp) kernel capturing different graph structures \cite{shervashidze2011weisfeiler}, PSCN algorithm that is based on graph CNN models \cite{niepert2016learning}, structure2vec models including DE-MF and DE-LBP \cite{dai2016discriminative} and DGCNN based on a new deep graph convolutional neural network architecture \cite{zhang2018end}.

\begin{figure}
	\centering
	\begin{subfigure}{.48\columnwidth}
		\centering
		\includegraphics[width=\linewidth]{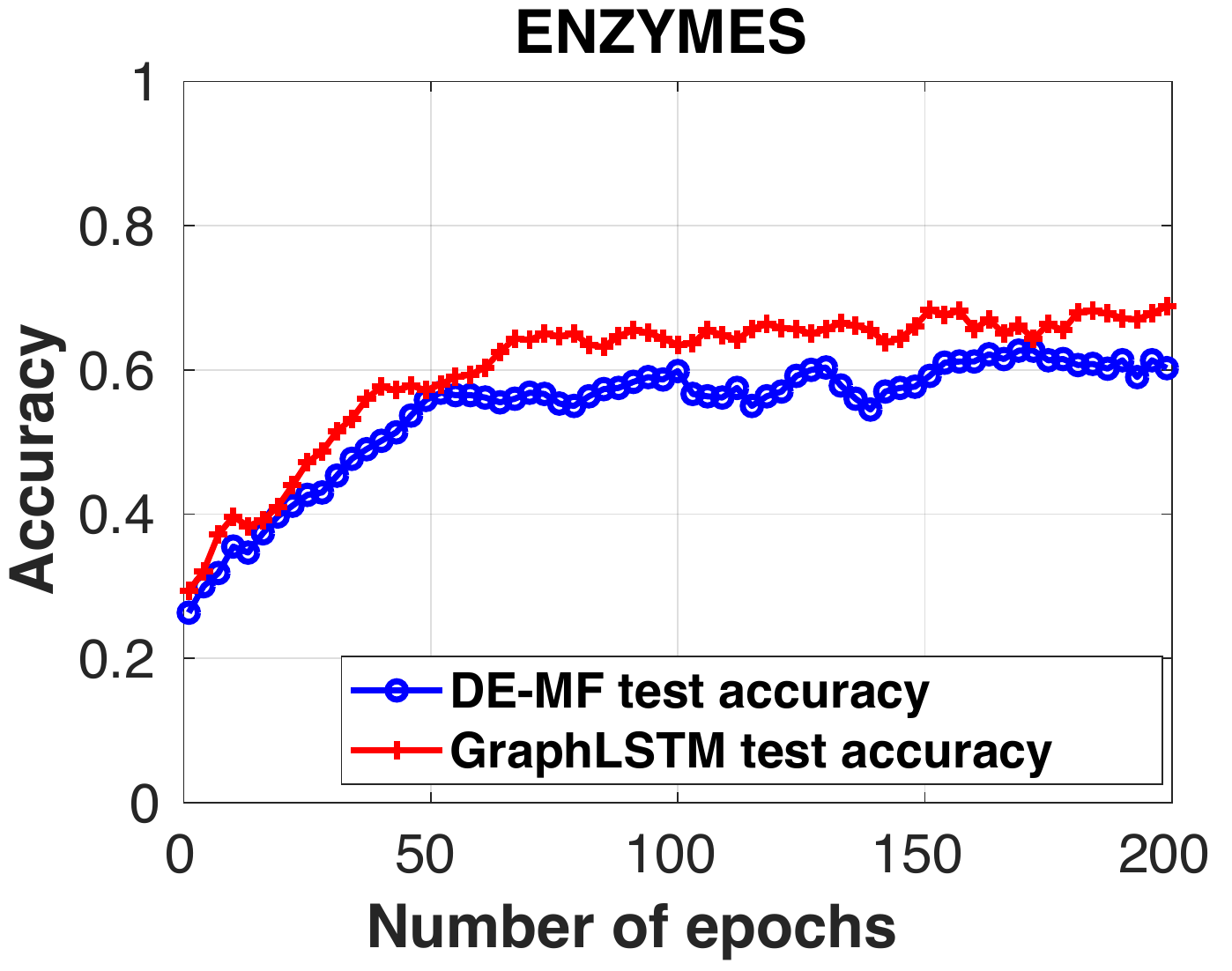}
		\label{enzymes_accuracy}
	\end{subfigure}
	\begin{subfigure}{.48\columnwidth}
		\centering
		\includegraphics[width=\linewidth]{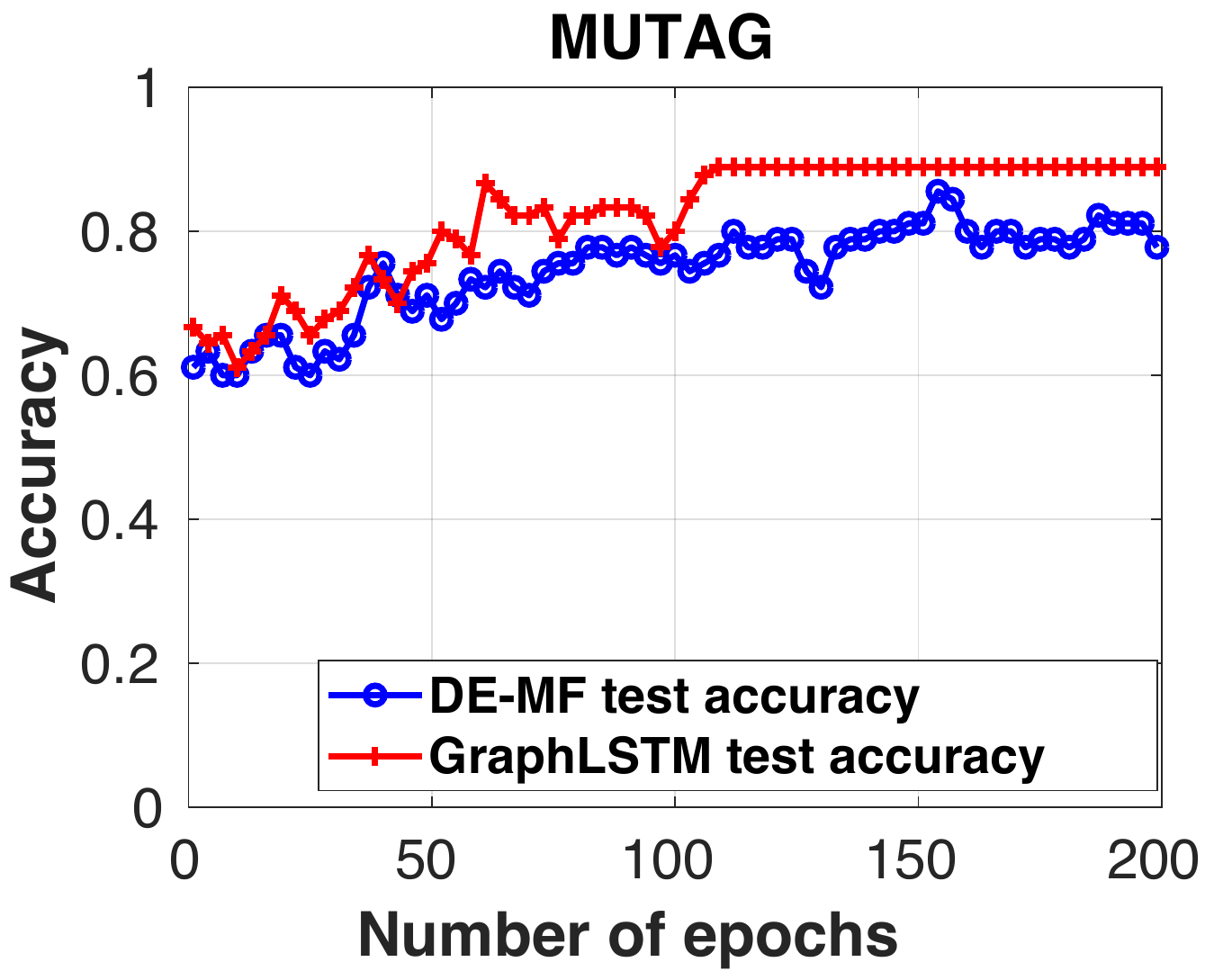}
		\label{mutag_accuracy}
	\end{subfigure}
	\caption{Classification accuracy on ENZYMES and MUTAG datasets with the number of epochs comparing GraphLSTM and DE-MF models. \cite{dai2016discriminative}.}
	\label{accuracy}
\end{figure}

\subsubsection{Results}
Table \ref{classification} shows the results of the classification accuracy on the five benchmark datasets. Our proposed model achieves comparable or superior results relative to the state-of-the art methods. 

Figure \ref{accuracy} further compares the classification accuracy of the GraphLSTM model and DE-MF model as a function of the number of epochs. Note that DE-MF model is a representation of the recent algorithms that tackle graph problems with convolutional neural network structures. For ENZYMES and MUTAG datasets, the GraphLSTM model achieves significantly better results in terms of the classification accuracy as well as convergence speed, which shows the superiority of RNN models for learning graph representations in the classification tasks.

\subsection{Discussion on the graph RNN model}
In this part, we discuss several important factors that affect the performance of our proposed graph RNN model.

\begin{figure}
	\centering
	\begin{subfigure}{.48\columnwidth}
		\centering
		\includegraphics[width=\linewidth]{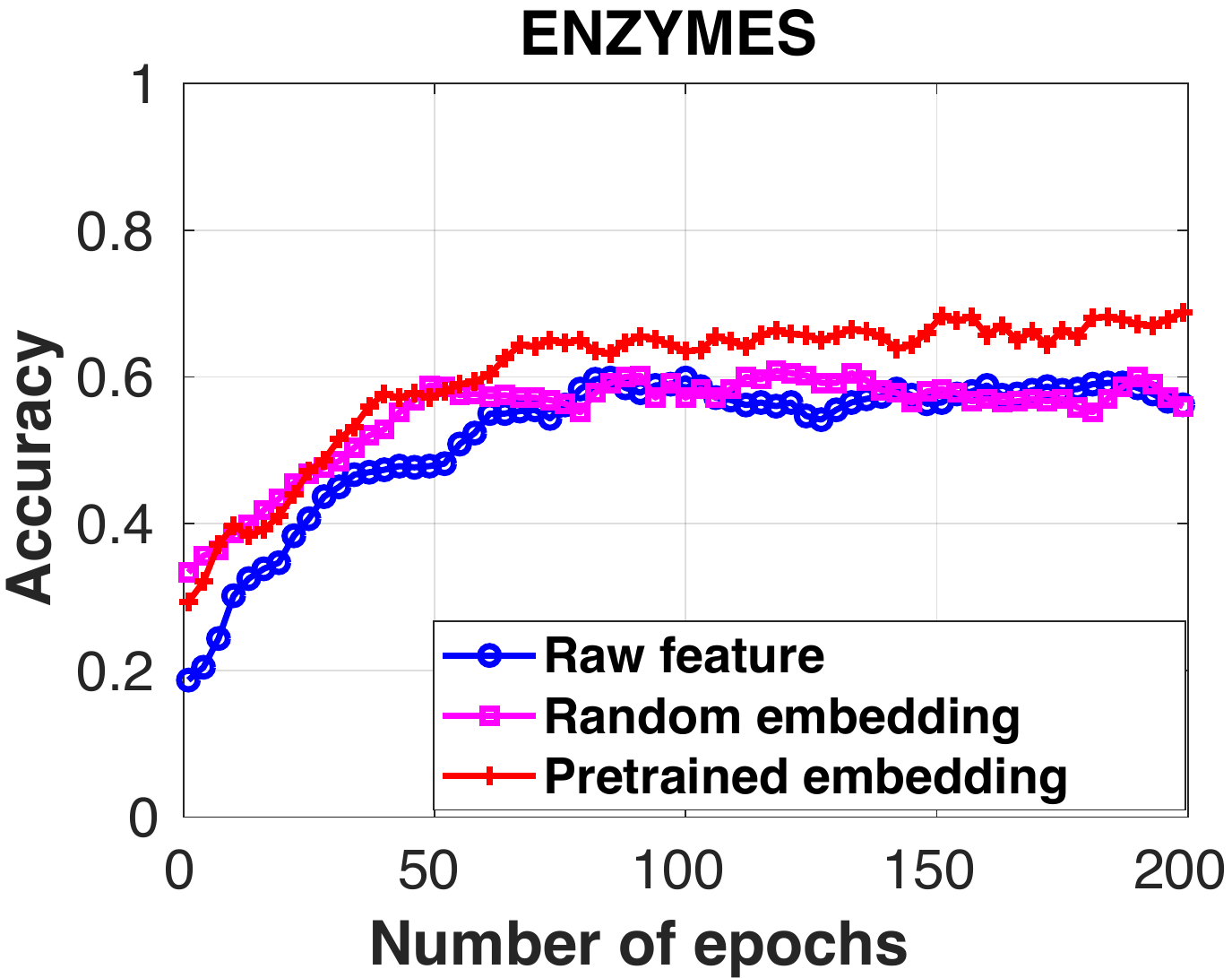}
		\label{enzymes_embedding}
	\end{subfigure}
	\begin{subfigure}{.5\columnwidth}
		\centering
		\includegraphics[width=\linewidth]{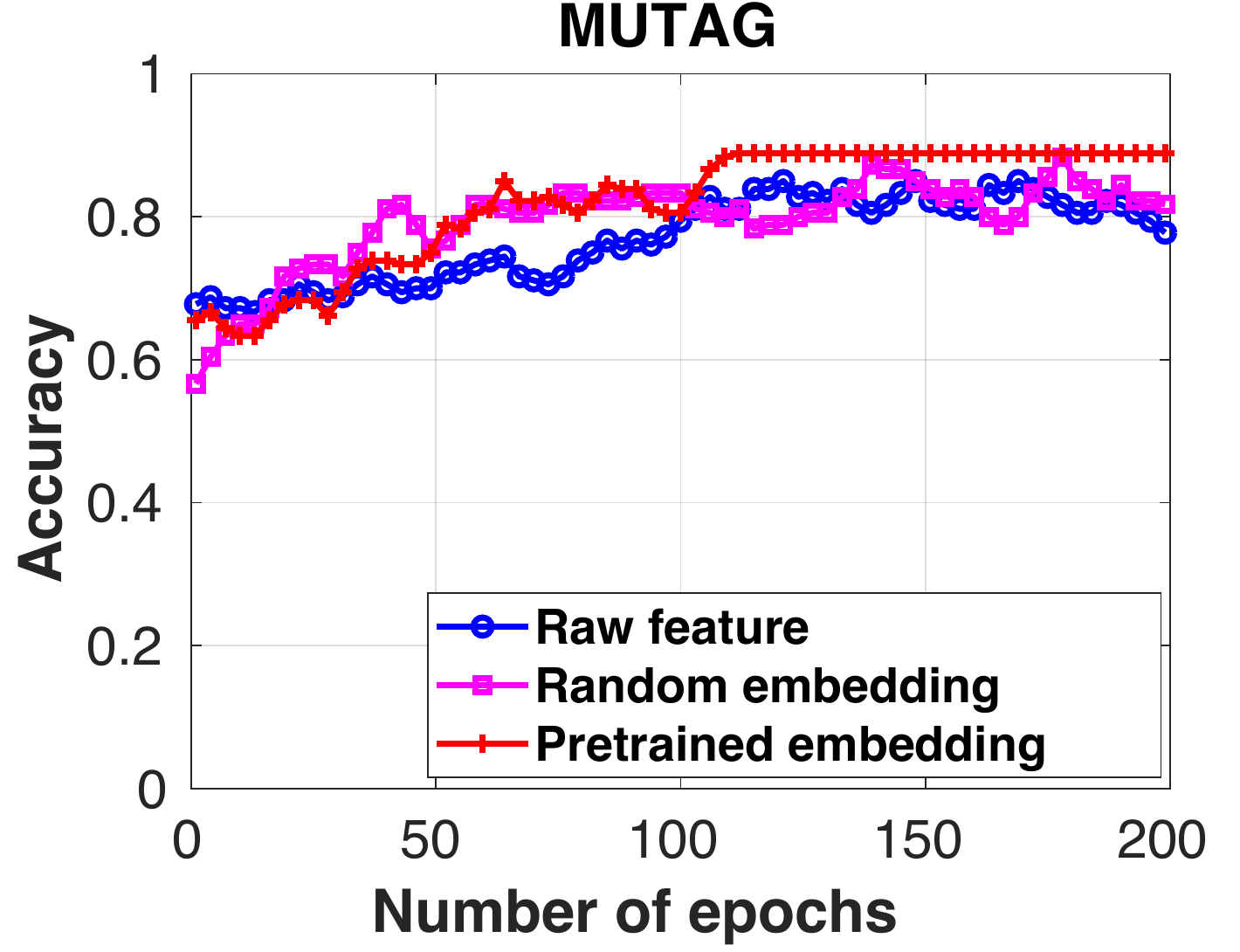}
		\label{mutag_embedding}
	\end{subfigure}
	\caption{Classification accuracy on benchmark datasets with different embedding methods.}
	\label{embedding}
\end{figure}

\subsubsection{Node representations}
Node embeddings, as the direct inputs on the RNN model, are important to the performance of the graph RNN models \cite{jin2016bag}. We evaluate the effect of different definitions of node representations on the performance of the GraphLSTM model. We compare the set of pretrained embeddings against the baseline node representations as briefly described below,
\begin{itemize}
	\item Raw features: the one-hot vectors $H\in \mathbb{R}^{n\times k}$ are directly used as the node representations.
	\item Randomized embedding: the embedding matrix $E \in \mathbb{R}^{k \times d}$ is randomly initialized as i.i.d. samples from the normal distribution with mean 0 and variance 1. 
\end{itemize}

Figure \ref{embedding} shows the results comparing the performance using the different definitions of node embeddings. The model with pretrained embeddings of graph nodes outperforms the other embeddings by a large margin. Hence the pretrained embeddings learned from graph structural information are effective for graph RNN models to learn graph representations. 

\begin{figure}[h]
	\centering
	\begin{subfigure}{.48\columnwidth}
		\centering
		\includegraphics[width=1\linewidth]{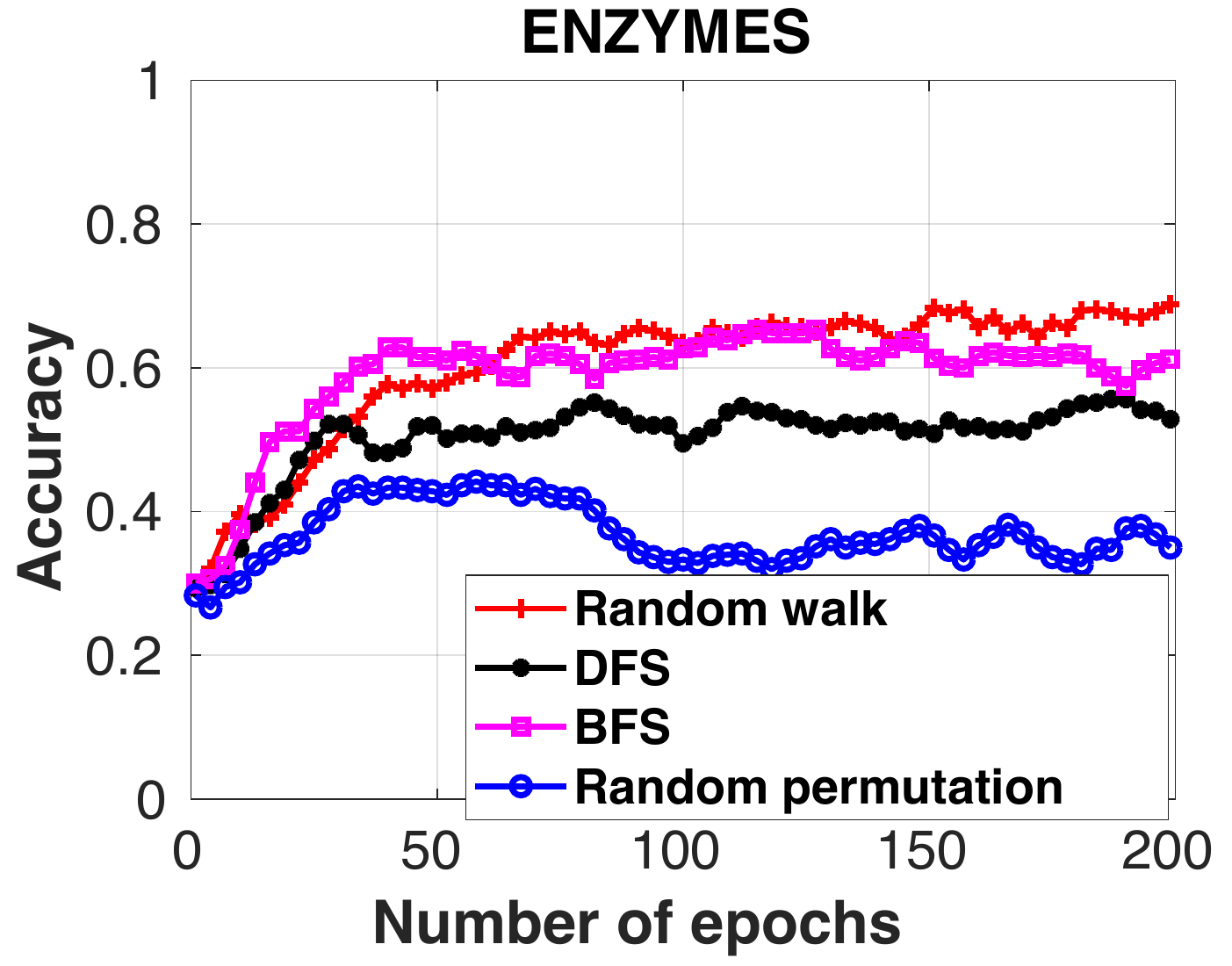}
		\label{enzymes_order}
	\end{subfigure}
	\begin{subfigure}{.48\columnwidth}
		\centering
		\includegraphics[width=\linewidth]{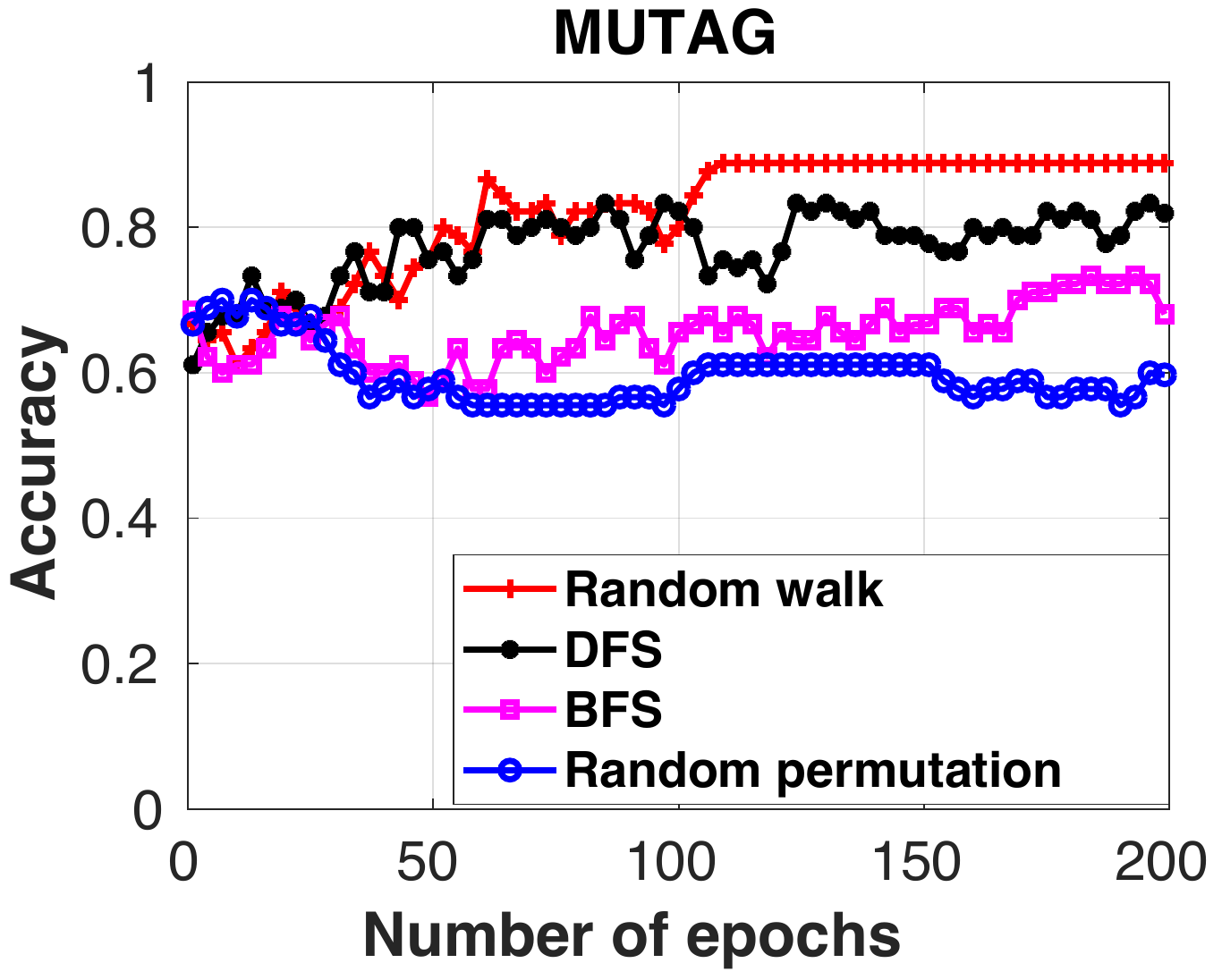}
		\label{mutag_order}
	\end{subfigure}
	\caption{Classification accuracy on benchmark datasets with different node ordering methods.}
	\label{order}
\end{figure}
\subsubsection{Node sequences}

We evaluate the impact of different node ordering methods on the prediction accuracy, which is shown in Figure \ref{order}. The baseline methods are the random permutation, Breadth First Search (BFS), and Depth First Search (DFS). Note that the baseline methods determine the graph sequences independent of the classification task. The results show that the order of the node sequence affects the classification accuracy by a large margin. The RNN model with the parameterized random walk approach achieves the best results compared to other graph ordering methods. The random permutations yields the worst performance which suggests preserving the proximity relationship of the original graph in the node sequence is important to learn the effective node representations in the classification tasks.

\section{Conclusion}
In this work, we propose a new graph learning model to directly learn graph-level representations with the recurrent neural network model from samples of variable-sized graphs. New node embedding methods are proposed to embed graph nodes into high-dimensional vector space which captures both the node features as well as the structural information. We propose a random walk approach with the Gumbel-Softmax approximation to generate continuous samples of node sequences with parameters learned from the classification objective. Empirical results show that our model outperforms the  state-of-the-art methods in graph classification. We also include a discussion of our model in terms of the node embedding and the order of node sequences, which illustrates the effectiveness of the strategies used.

\bibliographystyle{unsrt}
\bibliography{ref}

\begin{thebibliography}{10}

\bibitem{hamilton2017representation}
William~L Hamilton, Rex Ying, and Jure Leskovec.
\newblock Representation learning on graphs: Methods and applications.
\newblock {\em arXiv preprint arXiv:1709.05584}, 2017.

\bibitem{bronstein2016geometric}
Michael~M Bronstein, Joan Bruna, Yann LeCun, Arthur Szlam, and Pierre
  Vandergheynst.
\newblock Geometric deep learning: going beyond euclidean data.
\newblock {\em arXiv preprint arXiv:1611.08097}, 2016.

\bibitem{duvenaud2015convolutional}
David~K Duvenaud, Dougal Maclaurin, Jorge Iparraguirre, Rafael Bombarell,
  Timothy Hirzel, Al{\'a}n Aspuru-Guzik, and Ryan~P Adams.
\newblock Convolutional networks on graphs for learning molecular fingerprints.
\newblock In {\em Advances in neural information processing systems}, pages
  2224--2232, 2015.

\bibitem{dai2016discriminative}
Hanjun Dai, Bo~Dai, and Le~Song.
\newblock Discriminative embeddings of latent variable models for structured
  data.
\newblock In {\em International Conference on Machine Learning}, pages
  2702--2711, 2016.

\bibitem{gilmer2017neural}
Justin Gilmer, Samuel~S Schoenholz, Patrick~F Riley, Oriol Vinyals, and
  George~E Dahl.
\newblock Neural message passing for quantum chemistry.
\newblock {\em ICML}, 2017.

\bibitem{kipf2016semi}
Thomas~N Kipf and Max Welling.
\newblock Semi-supervised classification with graph convolutional networks.
\newblock {\em ICLR}, 2016.

\bibitem{khasanova2017graph}
Renata Khasanova and Pascal Frossard.
\newblock Graph-based isometry invariant representation learning.
\newblock {\em arXiv preprint arXiv:1703.00356}, 2017.

\bibitem{mikolov2013efficient}
Tomas Mikolov, Kai Chen, Greg Corrado, and Jeffrey Dean.
\newblock Efficient estimation of word representations in vector space.
\newblock {\em arXiv preprint arXiv:1301.3781}, 2013.

\bibitem{mikolov2013distributed}
Tomas Mikolov, Ilya Sutskever, Kai Chen, Greg~S Corrado, and Jeff Dean.
\newblock Distributed representations of words and phrases and their
  compositionality.
\newblock In {\em Advances in neural information processing systems}, pages
  3111--3119, 2013.

\bibitem{gartner2003graph}
Thomas G{\"a}rtner, Peter Flach, and Stefan Wrobel.
\newblock On graph kernels: Hardness results and efficient alternatives.
\newblock {\em Learning Theory and Kernel Machines}, pages 129--143, 2003.

\bibitem{kashima2003marginalized}
Hisashi Kashima, Koji Tsuda, and Akihiro Inokuchi.
\newblock Marginalized kernels between labeled graphs.
\newblock In {\em Proceedings of the 20th international conference on machine
  learning (ICML-03)}, pages 321--328, 2003.

\bibitem{horvath2004cyclic}
Tam{\'a}s Horv{\'a}th, Thomas G{\"a}rtner, and Stefan Wrobel.
\newblock Cyclic pattern kernels for predictive graph mining.
\newblock In {\em Proceedings of the tenth ACM SIGKDD international conference
  on Knowledge discovery and data mining}, pages 158--167. ACM, 2004.

\bibitem{douglas2011weisfeiler}
Brendan~L Douglas.
\newblock The weisfeiler-lehman method and graph isomorphism testing.
\newblock {\em arXiv preprint arXiv:1101.5211}, 2011.

\bibitem{defferrard2016convolutional}
Micha{\"e}l Defferrard, Xavier Bresson, and Pierre Vandergheynst.
\newblock Convolutional neural networks on graphs with fast localized spectral
  filtering.
\newblock In {\em Advances in Neural Information Processing Systems}, pages
  3844--3852, 2016.

\bibitem{niepert2016learning}
Mathias Niepert, Mohamed Ahmed, and Konstantin Kutzkov.
\newblock Learning convolutional neural networks for graphs.
\newblock In {\em International Conference on Machine Learning}, pages
  2014--2023, 2016.

\bibitem{li2015gated}
Yujia Li, Daniel Tarlow, Marc Brockschmidt, and Richard Zemel.
\newblock Gated graph sequence neural networks.
\newblock {\em arXiv preprint arXiv:1511.05493}, 2015.

\bibitem{cho2014learning}
Kyunghyun Cho, Bart Van~Merri{\"e}nboer, Caglar Gulcehre, Dzmitry Bahdanau,
  Fethi Bougares, Holger Schwenk, and Yoshua Bengio.
\newblock Learning phrase representations using rnn encoder-decoder for
  statistical machine translation.
\newblock {\em arXiv preprint arXiv:1406.1078}, 2014.

\bibitem{yin2017comparative}
Wenpeng Yin, Katharina Kann, Mo~Yu, and Hinrich Sch{\"u}tze.
\newblock Comparative study of cnn and rnn for natural language processing.
\newblock {\em arXiv preprint arXiv:1702.01923}, 2017.

\bibitem{jain2016structural}
Ashesh Jain, Amir~R Zamir, Silvio Savarese, and Ashutosh Saxena.
\newblock Structural-rnn: Deep learning on spatio-temporal graphs.
\newblock In {\em Proceedings of the IEEE Conference on Computer Vision and
  Pattern Recognition}, pages 5308--5317, 2016.

\bibitem{yuan2017temporal}
Yuan Yuan, Xiaodan Liang, Xiaolong Wang, Dit-Yan Yeung, and Abhinav Gupta.
\newblock Temporal dynamic graph lstm for action-driven video object detection.
\newblock In {\em ICCV}, pages 1819--1828, 2017.

\bibitem{you2018graphrnn}
Jiaxuan You, Rex Ying, Xiang Ren, William~L Hamilton, and Jure Leskovec.
\newblock Graphrnn: A deep generative model for graphs.
\newblock {\em International Conference on Machine Learning}, 2018.

\bibitem{grover2016node2vec}
Aditya Grover and Jure Leskovec.
\newblock node2vec: Scalable feature learning for networks.
\newblock In {\em Proceedings of the 22nd ACM SIGKDD international conference
  on Knowledge discovery and data mining}, pages 855--864. ACM, 2016.

\bibitem{hamilton2017inductive}
Will Hamilton, Zhitao Ying, and Jure Leskovec.
\newblock Inductive representation learning on large graphs.
\newblock In {\em Advances in Neural Information Processing Systems}, pages
  1025--1035, 2017.

\bibitem{perozzi2014deepwalk}
Bryan Perozzi, Rami Al-Rfou, and Steven Skiena.
\newblock Deepwalk: Online learning of social representations.
\newblock In {\em Proceedings of the 20th ACM SIGKDD international conference
  on Knowledge discovery and data mining}, pages 701--710. ACM, 2014.

\bibitem{vinyals2015order}
Oriol Vinyals, Samy Bengio, and Manjunath Kudlur.
\newblock Order matters: Sequence to sequence for sets.
\newblock {\em arXiv preprint arXiv:1511.06391}, 2015.

\bibitem{jang2016categorical}
Eric Jang, Shixiang Gu, and Ben Poole.
\newblock Categorical reparameterization with gumbel-softmax.
\newblock {\em arXiv preprint arXiv:1611.01144}, 2016.

\bibitem{maddison2016concrete}
Chris~J Maddison, Andriy Mnih, and Yee~Whye Teh.
\newblock The concrete distribution: A continuous relaxation of discrete random
  variables.
\newblock {\em arXiv preprint arXiv:1611.00712}, 2016.

\bibitem{sugiyama2015halting}
Mahito Sugiyama and Karsten Borgwardt.
\newblock Halting in random walk kernels.
\newblock In {\em Advances in neural information processing systems}, pages
  1639--1647, 2015.

\bibitem{liang1994classification}
Chengzhi Liang and Kurt Mislow.
\newblock Classification of topologically chiral molecules.
\newblock {\em Journal of Mathematical Chemistry}, 15(1):245--260, 1994.

\bibitem{shervashidze2011weisfeiler}
Nino Shervashidze, Pascal Schweitzer, Erik Jan~van Leeuwen, Kurt Mehlhorn, and
  Karsten~M Borgwardt.
\newblock Weisfeiler-lehman graph kernels.
\newblock {\em Journal of Machine Learning Research}, 12(Sep):2539--2561, 2011.

\bibitem{wale2008comparison}
Nikil Wale, Ian~A Watson, and George Karypis.
\newblock Comparison of descriptor spaces for chemical compound retrieval and
  classification.
\newblock {\em Knowledge and Information Systems}, 14(3):347--375, 2008.

\bibitem{borgwardt2005shortest}
Karsten~M Borgwardt and Hans-Peter Kriegel.
\newblock Shortest-path kernels on graphs.
\newblock In {\em Data Mining, Fifth IEEE International Conference on}, pages
  8--pp. IEEE, 2005.

\bibitem{kingma2014adam}
Diederik~P Kingma and Jimmy Ba.
\newblock Adam: A method for stochastic optimization.
\newblock {\em arXiv preprint arXiv:1412.6980}, 2014.

\bibitem{zhang2018end}
Muhan Zhang, Zhicheng Cui, Marion Neumann, and Yixin Chen.
\newblock An end-to-end deep learning architecture for graph classification.
\newblock In {\em Proceedings of AAAI Conference on Artificial Inteligence},
  2018.

\bibitem{jin2016bag}
Peng Jin, Yue Zhang, Xingyuan Chen, and Yunqing Xia.
\newblock Bag-of-embeddings for text classification.
\newblock In {\em IJCAI}, volume~16, pages 2824--2830, 2016.

\end{thebibliography}
\end{document}